\pdfoutput=1

\documentclass[11pt]{article}

\usepackage{acl}

\usepackage{times}
\usepackage{latexsym}

\usepackage[T1]{fontenc}

\usepackage[utf8]{inputenc}

\usepackage{microtype}

\usepackage[normalem]{ulem}
\useunder{\uline}{\ul}{}
\usepackage{multirow}
\usepackage{multicol}
\usepackage{graphicx}

\usepackage{amssymb}
\usepackage{amsmath}
\usepackage{enumitem}

\title{Effective Unsupervised Constrained Text Generation \\based on Perturbed Masking}

\author{Yingwen Fu$^{1,2}$\thanks{` Equal contribution. This work was conducted when Yingwen Fu was interning at NetEase Games AI Lab.}  , Wenjie Ou$^{2,*}$\thanks{` Corresponding author.} , Zhou Yu$^{3}$, Yue Lin$^{2}$\\
$^1$Guangdong University of Foreign Studies, Guangzhou, China \\ $^2$NetEase Games AI Lab, Guangzhou, China \\
$^3$Columbia University \\
\tt $^1$20201010002@gdufs.edu.cn, \tt $^3$zy2461@columbia.edu, \\ \tt $^2$\{ouwenjie, gzlinyue\}@corp.netease.com}

\begin{document}
\maketitle
\begin{abstract}

Unsupervised constrained text generation aims to generate text under a given set of constraints without any supervised data. Current state-of-the-art methods stochastically sample edit positions and actions, which may cause unnecessary search steps. In this paper, we propose PMCTG to improve effectiveness by searching for the best edit position and action in each step. Specifically, PMCTG extends perturbed masking technique to effectively search for the most incongruent token to edit. Then it introduces four multi-aspect scoring functions to select edit action to further reduce search difficulty. Since PMCTG does not require supervised data, it could be applied to different generation tasks. We show that under the unsupervised setting, PMCTG achieves new state-of-the-art results in two representative tasks, namely keywords-to-sentence generation and paraphrasing.
\end{abstract}

\section{Introduction}
Constrained text generation is the task of generating text that satisfies a given set of constraints, and it serves many real-world text generation applications, such as dialogue generation \cite{DBLP:conf/naacl/LiGBGD16} and summarization \cite{DBLP:conf/acl/SeeLM17}. There are broadly two types of constraints: Hard constraints such as including a set of given words or phrases in the generated text. Example 1 in Table \ref{table:1} shows that the keywords “\textit{You}” and “\textit{beautiful}” must occur in the generated sentence. Soft constraints such as acquiring the generated text to be semantically similar to the original text. Example 2 in Table \ref{table:1} shows a pair of paraphrases where “\textit{What are the effective ways to learn cs?}” and “\textit{How to learn cs effectively?}” share a similar meaning.

Conventional approaches model the task in an encoding-decoding paradigm with a supervised setting \cite{DBLP:conf/coling/PrakashHLDQLF16,DBLP:conf/aaai/GuptaASR18}. However, these methods have certain shortcomings for two constrained generation tasks. For hard constrained text generation, without external constrained means, these methods are difficult to guarantee that the generated text can satisfy all constraints. For soft constrained one, conventional methods treat it as a machine translation (MT) task \cite{DBLP:conf/nips/SutskeverVL14} and require massive parallel supervised data for training. Unfortunately, constructing such datasets is resource-intensive. 
In addition, domain-specific supervised models may be difficult to transfer to new domains. \cite{DBLP:conf/acl/LiJSL19}.

\begin{table}[]
\centering
\begin{tabular}{l|l|l}
\hline
  \textbf{No.} &
  \textbf{Original Text} &
  \textbf{Generated Text} \\ \hline
  \begin{tabular}[c]{@{}l@{}}1\end{tabular} &
  \begin{tabular}[c]{@{}l@{}}You,\\ beautiful\end{tabular} &
  You are so beautiful . \\ \hline
  \begin{tabular}[c]{@{}l@{}}2\end{tabular} &
  \begin{tabular}[c]{@{}l@{}}How to learn\\ cs effectively?\end{tabular} &
  \begin{tabular}[c]{@{}l@{}}What are the effective\\ ways to learn cs?\end{tabular} \\ \hline
\end{tabular}
\caption{Examples on constrained text generation.}
\label{table:1}
\end{table}

Recently, unsupervised text generation is proposed to address the above challenges. There are mainly two research directions: Beam search-based method aims to generate candidates in order from left to right that satisfy the constraints in each step, inspired by MT methods \cite{DBLP:conf/acl/HokampL17, DBLP:conf/naacl/PostV18}. However, the search space of MT systems is relatively small, while when applied to other generation tasks, such as paraphrase, this approach does not work as optimally as expected because of a much larger search space \cite{DBLP:conf/emnlp/Sha20}. Local edit-based method represented by CGMH \cite{DBLP:conf/aaai/MiaoZMYL19} and USPA \cite{DBLP:conf/acl/LiuMMZZS20} is another effective solution. These methods propose stochastic local edit strategies to search for reasonable sentences in a huge search space based on the given constraints. One main concern is that these methods may take a long time to search for the optimal solution because they are based on stochastic strategies. Intuitively, they need more search steps to converge. G2LC \cite{DBLP:conf/emnlp/Sha20} utilizes gradients to determine edit positions and actions to improve search effectiveness. But it still relies on supervised data.

Dedicated to improving the local edit-based methods, in this paper, we propose a framework PMCTG (\textbf{P}erturbed \textbf{M}asking for \textbf{C}onstrained \textbf{T}ext \textbf{G}eneration) for constrained text generation. PMCTG focuses on controlling the search direction and reducing the search steps by searching for the best edit position and action at each step. Specifically, PMCTG extends perturbed masking \cite{DBLP:conf/acl/WuCKL20} from a pre-trained BERT model \cite{DBLP:conf/naacl/DevlinCLT19} to find the best edit position in the sequence. Perturbed masking aims to estimate the correlation between tokens in a sequence, which can be naturally used to find the edit location. We also propose a series of scoring functions for different tasks to select the edit action. PMCTG does not rely on supervised data and only needs a pre-trained BERT model to perform perturbed masking.

We evaluate PMCTG in two constrained text generation tasks, namely keywords-to-sentence generation and paraphrasing. Experimental results show that PMCTG tends to achieve new state-of-the-art performance over multiple baselines. In summary, the contributions are as follows: 
\begin{enumerate}[itemsep=2pt,topsep=0pt,parsep=0pt]
    \item We extend perturbed masking to constrained text generation which can find edit positions more effectively. 
    \item We design different scoring functions to select the best action effectively. With different scoring functions, PMCTG can be extended to various generation tasks \cite{DBLP:conf/emnlp/KikuchiNSTO16,ficler2017controlling,DBLP:conf/icml/HuYLSX17}.
    \item We demonstrate our method's state-of-the-art performance in keywords-to-sentence generation and paraphrasing tasks.
\end{enumerate}

\section{Related Work}
\subsection{Constrained Text Generation}
Constrained text generation is formulated as a supervised sequence-to-sequence problem under the encoding-decoding paradigm \cite{DBLP:conf/nips/SutskeverVL14}. For example, \cite{DBLP:conf/coling/PrakashHLDQLF16} and \cite{DBLP:conf/acl/LiJSL19} respectively propose a stacked residual LSTM network and a transformer-based model \cite{DBLP:conf/nips/VaswaniSPUJGKP17}, and \cite{DBLP:conf/aaai/GuptaASR18} propose to leverage a combination of variational autoencoders (VAEs) with LSTM models to generate paraphrases. A new sentence generation model is proposed by \cite{DBLP:journals/tacl/GuuHOL18}, where a prototype sentence is first extracted from the training corpus and then edited into a new sentence. However, these methods do not support constraint integration  \cite{DBLP:conf/aaai/MiaoZMYL19}. Later, some works have attempted to add constraints to the generated models. \cite{DBLP:conf/acl/WuebkerGDHL16} and \cite{DBLP:conf/amta/KnowlesK16} utilize prefixes to guide the target text generation. \cite{DBLP:conf/coling/MouSYL0J16} use pointwise mutual information (PMI) to predict a keyword and treat it as a constraint to generate target text. However, these methods always bind the constraints to the original model and are therefore difficult to apply to new domains and new generation models \cite{DBLP:conf/acl/LiJSL19}. Moreover, the above approaches rely on an adequate parallel supervised corpus, which is hard to obtain in real-world application scenarios.

Unsupervised constrained text generation has become a research hotspot due to its low training cost and mitigation of insufficient training data. VAEs and their variants \cite{DBLP:conf/conll/BowmanVVDJB16,DBLP:conf/acl/RoyG19} are leveraged to generate sentences from a continuous latent space. These methods can effectively get rid of the reliance on supervised datasets but remain difficult to control and incorporate generative constraints.

Beam search is a representative approach for unsupervised constrained text generation. Grid Beam Search (GBS) \cite{DBLP:conf/acl/HokampL17} is an algorithm that extends beam search by allowing the inclusion of pre-specified lexical constraints. \cite{DBLP:conf/naacl/PostV18} propose Dynamic Beam Allocation (DBA), a much faster beam search-based method with hard lexical constraints. \cite{DBLP:conf/emnlp/ZhangWLGBD20} propose an insertion-based approach consisting of insertion-based generative pre-training and inner-layer beam search. For the tasks where the search space is limited (represented by machine translation), these methods work well. However, when faced with a large search space, they do not work as optimally as expected \cite{DBLP:conf/emnlp/Sha20}.

Local edit-based methods have attracted attention recently, as they can help to reduce search spaces. CGMH \cite{DBLP:conf/aaai/MiaoZMYL19} applies the Metropolis-Hastings algorithm \cite{metropolis1953equation} to unsupervised constrained generation. UPSA \cite{DBLP:conf/acl/LiuMMZZS20} is another local edit-based method. It directly models paraphrasing as an optimization problem and uses simulated annealing to solve it. However, these models may require many steps and running time to generate reasonable sentences since they are based on stochastic strategies. \cite{DBLP:conf/emnlp/Sha20} proposes a gradient-guided method G2LC that uses token gradients to determine the edit actions and positions, making the generation process more controllable. However, a problem with G2LC is that it still relies on the supervised corpus to train a binary classification model to serve their semantic similarity objective.

\subsection{Perturbed Masking}
Perturbed masking \cite{DBLP:conf/acl/WuCKL20} is a parameter-free probing technique to analyze and interpret pre-trained models. Based on a pre-trained BERT-based model with masked language modeling (MLM) objective, it can measure the impact a token has on predicting another token. It is originally used in syntax-based tasks such as syntactic parsing and discourse dependency parsing. 

In this paper, we extend perturbed masking to constrained text generation. For the edit-based approach edits only one token at each step, we need to find the token with the highest incongruency to edit. Our insight is to use perturbed masking to present the congruency between different tokens. We believe that the token with the weakest correlation with its adjacent tokens has the highest incongruency and thus it is the most probable to edit. Perturbed masking can evaluate the impact of one token on another and a high impact factor means that the token has a high impact on its adjacent tokens and we consider these chunks (the current token with its adjacent tokens) are congruent. Therefore, we can edit the tokens in chunks with low impact to make these chunks more congruent.

\section{Methodology}

In this section, we would introduce the proposed PMCTG by first introducing the specific process of using perturbed masking to select edit positions, and then explaining the proposed scoring functions and the use of them to select the edit actions.

\subsection{Edit Position Selection}

Most previous works select edit locations stochastically, which lead to many unnecessary search steps. To reduce the search steps, we propose to use perturbed masking \cite{DBLP:conf/acl/WuCKL20} to sample the edit position. 

\noindent \textbf{Background.} Perturbed masking technique is proposed to assess the inter-token information (i.e., the impact one token has on another token in a sequence) based on masked language modeling (MLM). It is originally used for dependency parsing.

Formally, given a sequence with $n$ tokens $\boldsymbol{x}=\{x_i\}^n_{i=1}$ and a pre-trained BERT-based model \cite{DBLP:conf/naacl/DevlinCLT19}  trained with MLM objective, we obtain contextual representations for each token $H(\boldsymbol{x})_i$. To quantify the impact a token $x_j$ has on another token $x_i$, we conduct the following three-step calculation:
\begin{enumerate}[itemsep=2pt,topsep=0pt,parsep=0pt]
    \item Replace $x_i$ with $[MASK]$ token and feed the new sequence $\boldsymbol{x}\backslash\{x_i\}$ into BERT, a contextual representation denoted as $H(\boldsymbol{x}\backslash\{x_i\})_i$ for $x_i$ is obtained.
    \item Replace $x_i$ and $x_j$ with $[MASK]$ token and feed the new sequence $\boldsymbol{x}\backslash\{x_i,x_j\}$ into BERT, another contextual representation denoted as $H(\boldsymbol{x}\backslash\{x_i,x_j\})_i$ for $x_i$ is obtained.
    \item Given a distance metric $d(,)$, compute the difference between two vectors $I(\boldsymbol{x}|x_j,x_i)=d(H(\boldsymbol{x}\backslash\{x_i\})_i, H(\boldsymbol{x}\backslash\{x_i,x_j\})_i)$. Euclidean distance is leveraged in this paper.
\end{enumerate}

$I(\boldsymbol{x}|x_j,x_i)$ indicates the impact $x_j$ has on $x_i$, where a higher value indicates a high impact, and vice versa. Intuitively, if $H(\boldsymbol{x}\backslash\{x_i\})_i$ and $H(\boldsymbol{x}\backslash\{x_i,x_j\})_i$ are similar, it means that the presence or absence of $x_j$ has little effect on the prediction of $x_i$, thus reflecting the low importance of $x_j$ to $x_i$.

\noindent \textbf{Position Selection.} It is natural to apply perturbed masking to select the edit position for constrained text generation. Based on perturbed masking technique, we compute the edit score for each token in the sequence and then sample the token with the highest score to edit. The token with minimal impact on its adjacent tokens indicates that it has the weakest correlation with adjacent tokens and therefore requires edit. We add the special tokens $[CLS]$ and $[SEP]$ to the original sentence and then use the pre-trained BERT to calculate the edit score for each token:

\begin{equation}
\begin{split}
    ES_i=1-\frac{1}{2}(I(\boldsymbol{x}|x_i,x_{i+1})+ I(\boldsymbol{x}|x_i,x_{i-1})) \label{eq1}
\end{split}
\end{equation}

Then we can get an edit score vector $\boldsymbol{ES}=\{ES_i\}^{n}_{i=0}$. Later, we feed it into a softmax layer and obtain the edit probabilities:

\begin{equation}
    p^{edit}_i = \frac{exp(ES_{i})}{\sum_{j}exp(ES_{j})}  \label{eq_softmax}
\end{equation}

After that, the $\boldsymbol{p^{edit}}$ is utilized as the weights to sample the edit position $x_e$ in $\boldsymbol{x}$ where $e$ indicates the edit position index. 

\subsection{Edit Action Selection}

After sampling the edit position, next we need to determine the edit action. The three edit actions we focus on are: insert, replace and delete. Specifically, our strategy in this step is to pre-implement the three actions first and then sample the actions based on their action scores. When scoring insertion action, we simply make the equal probability of the front or back of the position for token insertion. We first introduce the scoring functions for different tasks and then explain the edit action selection based on the action scores.

\subsubsection{Scoring Function Design}
We propose multiple scoring functions to improve the generated text. Given the initial sentence $\boldsymbol{x_0}$ with $n$ tokens and the generated sentence $\boldsymbol{x_*}$ with $m$ tokens, the scoring functions include fluency, editorial rationality, semantic similarity, and diversity.

\noindent \textbf{Fluency.} The primary condition for a reasonable sentence is fluency, thus we use the average negative log-likelihood to estimate a sentence's fluency based on a forward language model. The score is calculated as: 

\begin{small}
\begin{equation}
    S_{flu}(\boldsymbol{x_*})=-\frac{1}{m}\sum^m_{i=1}logp_{LM}(x_{*,i}|x_{*,<i}) \label{eq2}
\end{equation}
\end{small}

\noindent \textbf{Editorial Rationality.} Since the sentence generation process is based on local edits, we further use perturbed masking to design a local edit score for different actions to evaluate their rationality. After a replacement action is executed at index $i$ in $\boldsymbol{x_0}$, we obtain the sentence $\boldsymbol{x_*}=\{x_{0,1}, x_{0,2}, …x_{0,i-1}, x', x_{0,i+1},…,x_{0,n}\}$, where $x'$ is the replaced token and $m=n$. Then we define the edit score as:

\begin{small}
\begin{equation}
\begin{split}
    S_{edit}(\boldsymbol{x_*})=\frac{1}{2}(I(\boldsymbol{x_*}|x',x_{0,i+1})+I(\boldsymbol{x_*}|x',x_{0,i-1})) \label{eq3}
\end{split}
\end{equation}
\end{small}

Similarly, after an insertion action, we obtain $\boldsymbol{x_*}=\{x_{0,1}, x_{0,2}, …x_{0,i}, x', x_{0,i+1},…,x_{0,n}\}$, where $x'$ is the inserted token and $m=n+1$. The edit score is calculated as:

\begin{small}
\begin{equation}
\begin{split}
    S_{edit}(\boldsymbol{x_*})=\frac{1}{2}(I(\boldsymbol{x_*}|x',x_{0,i+1})+I(\boldsymbol{x_*}|x',x_{0,i})) \label{eq4}
\end{split}
\end{equation}
\end{small}

After a deletion action, we obtain $\boldsymbol{x_*}=\{x_{0,1}, x_{0,2}, …x_{0,i-1}, x_{0,i+1},…,x_{0,n}\}$, where $m=n-1$. The edit score calculated for deletion is a little different from replacement and insertion action:

\begin{small}
\begin{equation}
\begin{split}
    S_{edit}(\boldsymbol{x_*})=\frac{1}{2}(I(\boldsymbol{x_*}|x_{0,i-1},x_{0,i+1})+ \\
    I(\boldsymbol{x_*}|x_{0,i+1},x_{0,i-1})) \label{eq5}
\end{split}
\end{equation}
\end{small}

\noindent \textbf{Semantic Similarity.} The semantic similarity consists of keyword similarity and sentence similarity. We use KeyBERT \cite{grootendorst2020keybert} to extract the keyword set $K$ from $\boldsymbol{x_0}$. And the pre-trained BERT is leveraged to encode $\boldsymbol{x_0}$ and $\boldsymbol{x_*}$, where $ik=idx(k)$ indicates the index of keyword $k$ in $\boldsymbol{x_0}$. The keyword similarity is defined as finding the closest token in $\boldsymbol{x_*}$ by computing their cosine similarity:

\begin{small}
\begin{equation}
\begin{split}
    & S_{sem,key}(\boldsymbol{x_*},\boldsymbol{x_0})= \\ & \frac{1}{|K|}\sum_{k\in K}\underset{i}{max}(cos(H(\boldsymbol{x_0})_{ik},H(\boldsymbol{x_*})_i)) \label{eq6}
\end{split}
\end{equation}
\end{small}

As for the sentence similarity, assuming that $H(x)$ indicates the $[CLS]$ representation in $x$ from BERT and is leveraged to present the whole sentence \cite{DBLP:conf/naacl/DevlinCLT19}, we define the sentence similarity $S_{sem,sen}(\boldsymbol{x_*,x_0})$ as:

\begin{small}
\begin{equation}
    S_{sem,sen}(\boldsymbol{x_*},\boldsymbol{x_0})=cos(H(\boldsymbol{x_0}),H(\boldsymbol{x_*})) \label{eq7}
\end{equation}
\end{small}

Altogether, the semantic similarity score is:

\begin{small}
\begin{equation}
\begin{split}
    S_{sem}(\boldsymbol{x_*},\boldsymbol{x_0})=S_{sem,key}(\boldsymbol{x_*},\boldsymbol{x_0})+S_{sem,sen}(\boldsymbol{x_*},\boldsymbol{x_0}) \label{eq8}
\end{split}
\end{equation}
\end{small}

\noindent \textbf{Diversity.} Followed \cite{DBLP:conf/acl/LiuMMZZS20}, a BLEU-based \cite{DBLP:conf/acl/PapineniRWZ02} function is adopted to evaluate the expression diversity of the original and generated sentence.

\begin{small}
\begin{equation}
    S_{exp}(\boldsymbol{x_*},\boldsymbol{x_0})=(1-BLEU(\boldsymbol{x_*},\boldsymbol{x_0})) \label{eq9}
\end{equation}
\end{small}

\subsubsection{Action Scoring}
As mentioned above, after sampling the edit position $i$, we need to determine the edit action by re-implementing three actions and sampling the actions based on their action scores. We generate the inserted and replaced candidate $x'$ from a language model such as LSTM \cite{hochreiter1997long} and GPT \cite{radford2019language}.

\begin{small}
\begin{equation}
    p_{candidate}=p_{LM}(x_{0,i}|x_{0,<i}) \label{eq10}
\end{equation}
\end{small}

We use $p_{candidate}$ as weights to sample $x'$.After obtaining the edit position $i$ and candidate $x'$, we need to calculate the edit score for each action. We adopt $S_{flu}$ and $S_{edit}$ the our scoring function for keywords-to-sentence generation:

\begin{small}
\begin{equation}
    S_{hard}(\boldsymbol{x_*})=\lambda_{flu}S_{flu}+\lambda_{edit}S_{edit} \label{eq11}
\end{equation}
\end{small}

and $S_{flu}$, $S_{sem}$, $S_{exp}$ and $S_{edit}$ for paraphrasing:

\begin{small}
\begin{equation}
\begin{split}
    S_{soft}(\boldsymbol{x_*})=& \lambda_{flu}S_{flu}+\lambda_{edit}S_{edit}+ \\ & \lambda_{sem}S_{sem}+\lambda_{exp}S_{exp} \label{eq12}
\end{split}
\end{equation}
\end{small}

Notably, since different scores are in different magnitudes, they need to be normalized to avoid the dominance of one specific score. After scoring different actions, we use the scores as weights to sample the edit action.

\subsection{Overall Searching Process}
With $\boldsymbol{x_0}$  (given keywords in the keywords-to-sentence generation task or original sentence in the paraphrasing task) as input, we repeat the above steps including edit position selection with perturbed masking and edit action selection with scoring functions for local edit. Until the maximum searching steps, we choose the sentence that achieves the highest score as the final output, according to (\ref{eq11}) for keywords-to-sentence generation task or (\ref{eq12}) for paraphrasing task respectively.

\section{Experiments}
We evaluate our method on two constrained text generation tasks, namely keywords-to-sentence generation and paraphrasing.

\subsection{Keywords-to-sentence Generation}
\noindent \textbf{Experimental Setting.} Keywords-to-sentence generation aims to generate a sentence containing the given keywords which is a representative hard constrained text generation task. We conduct keywords-to-sentence generation experiments on the One-Billion-token dataset\footnote{http://www.statmt.org/lm-benchmark/} \cite{DBLP:conf/interspeech/ChelbaMSGBKR14}. Two language models for generation, namely two-layer LSTM (followed as \cite{DBLP:conf/aaai/MiaoZMYL19,DBLP:conf/emnlp/Sha20}) and GPT \cite{radford2019language}, are evaluated. Following \cite{DBLP:conf/acl/GururanganMSLBD20}, in order to adapt the language models to the specific domain, we randomly sample 5 million sentences to continually pre-train BERT-based-cased\footnote{https://huggingface.co/bert-base-cased} and GPT2\footnote{https://huggingface.co/gpt2}. 3 thousand sentences are held out as the test set. 

As for hyperparameters, for each test sentence, we randomly sample 1 to 4 keywords as hard constraints. Following previous works \cite{DBLP:conf/aaai/MiaoZMYL19,DBLP:conf/emnlp/Sha20}, the initial sentence for searching is the concatenation of the keywords. The maximum searching step set in this task is 100. And $\lambda_{flu}$ and $\lambda_{edit}$ are set as $1$ in equation (12). Besides, when the keyword indexes are sampled as edit positions, we directly conduct insert action since the keywords cannot be replaced and deleted.

As for evaluation metrics, the generated target sentence is measured by negative log-likelihood (NLL) loss. NLL is given by a third-party language model which is an n-gram Kneser-Ney language model \cite{DBLP:conf/wmt/Heafield11} trained in a monolingual English corpus from WMT18\footnote{http://www.statmt.org/wmt18/translation-task.html}. In addition to automatic evaluation metrics, we also introduce human evaluation. Specifically, we invite 3 experts who are fluent English speakers to score the generated sentences according to their quality. The score ranges from 0 to 1 with an accuracy of two decimal places, where 1 indicates the best score. The automatic and human evaluation criteria are consistent with previous works \cite{DBLP:conf/emnlp/Sha20}. The scoring guideline is shown in Table \ref{table:2}.

\begin{table}[]
\centering
\begin{tabular}{l|l}
\hline
\textbf{Score}                                                     & \textbf{Description}                                                                                                      \\ \hline
1.00                                                              & \begin{tabular}[c]{@{}l@{}}Completely fluent.\end{tabular}                                           \\ \hline
0.75                                                              & \begin{tabular}[c]{@{}l@{}}Generally fluent with a few \\ grammatical errors.\end{tabular}                                           \\ \hline
0.50                                                              & \begin{tabular}[c]{@{}l@{}}Generally fluent with many \\ grammatical errors.\end{tabular}                                           \\ \hline
0.25                                                              & \begin{tabular}[c]{@{}l@{}}The whole sentence are not fluent,\\ but parts of it do.\end{tabular}                                           \\ \hline
0.00                                                              & \begin{tabular}[c]{@{}l@{}}Not readable.\end{tabular}                                           \\ \hline

\end{tabular}
\caption{Fluency scoring guideline.}
\label{table:2}
\end{table}

\begin{table*}[]
\centering
\begin{tabular}{l|ccccc|ccccc}
\hline
\multirow{2}{*}{\textbf{Models}} & \multicolumn{5}{c|}{\textbf{NLL}} & \multicolumn{5}{c}{\textbf{Score (Human Evaluation)}}       \\\cline{2-11} 
         & 1    & 2    & 3    & 4  & avg    & 1    & 2    & 3    & 4 & avg    \\ \hline
\textbf{seq-B/F}  & 7.80  & / & / & / & /    & 0.11 & / & / & / & /    \\
\textbf{asyn-B/F} & 8.30  & / & / & / & /    & 0.09 & / & / & / & /    \\
\textbf{GBS}      & 7.42 & 8.72 & 8.59 & 9.63 & 8.59 & 0.32 & 0.55 & 0.49 & 0.55 & 0.48 \\
\textbf{DBA}      & 7.41 & 8.58 & 8.54 & 9.25 & 8.45 & 0.43 & 0.53 & 0.54 & 0.59 & 0.52 \\
\textbf{CGMH}     & 7.04 & 7.57 & 8.26 & 7.92 & 7.70 & 0.45 & 0.61 & 0.56 & 0.65 & 0.57 \\
\textbf{G2LC}     & 7.02 & 7.46 & 8.01 & 7.76 & 7.56 & 0.47 & \textbf{0.73} & 0.65 & 0.67 & 0.63 \\ \hline
\textbf{PMCTG-GPT2}  & 6.98  & 7.45 & \textbf{7.69} & 7.89 & 7.50  & 0.51 & 0.68 & \textbf{0.70} & 0.72  & 0.65 \\
\textbf{PMCTG-LSTM}  & \textbf{6.92} & \textbf{7.33} & 7.93 & \textbf{7.68} & \textbf{7.47} & \textbf{0.53} & 0.69 & 0.68 & \textbf{0.74} & \textbf{0.66} \\ \hline
\end{tabular}
\caption{Performance on keywords-to-sentence generation task. Lower NLL and higher score indicate better result. 1,2,3 and 4 present the keyword numbers and avg indicates the average score. }
\label{table:3}
\end{table*}

\noindent \textbf{Baseline.} We compare our method with several advanced methods:
\begin{itemize}
\setlength{\itemsep}{0pt}
\setlength{\parsep}{0pt}
\setlength{\parskip}{0pt}
    \item \textbf{sep-B/F} \cite{DBLP:conf/coling/MouSYL0J16} is a variant of the backward forward model. In sep-B/F, the backward and forward sequences respectively behind and after the keyword are generated separately. It supports only one keyword.
    \item \textbf{asyn-B/F} \cite{DBLP:conf/coling/MouSYL0J16} is similar to sep-B/F. The difference is that the two sequences are generated asynchronously, i.e., the backward sequence is first generated, and then the forward sequence is generated based on the backward one.
    \item \textbf{GBS} \cite{DBLP:conf/acl/HokampL17} is a searching approach that aims to search for a valid solution in the constrained search space of the generator with grid beam search.
    \item \textbf{DBA} \cite{DBLP:conf/naacl/PostV18} is another beam search-based approach with a higher search speed.
    \item \textbf{CGMH} \cite{DBLP:conf/aaai/MiaoZMYL19} is a stochastic search method based on Metropolis-Hastings sampling.
    \item \textbf{G2LC} \cite{DBLP:conf/emnlp/Sha20} is a gradient-guided approach. It improves CGMH by leveraging gradient to decide the edit positions and actions.
\end{itemize}

\noindent \textbf{Automatic and Human Evaluation Results.} Table \ref{table:3} shows the performance of multiple methods on keywords-to-sentence generation task. Among different kinds of methods, we can see that the local edit-based methods work better than beam search-based methods, indicating their superior searching ability. CGMH can narrow the search space and make it easy to find higher-quality sentences. G2LC and PMCTG outperform CGMH, which illustrates the importance of determining the correct edit position and action for each step. Exploration and strategies for these two issues can better guide the model to find a more optimal solution, while also greatly reducing the waste of potentially non-essential search steps. Overall, the proposed PMCTG model outperforms other methods on average in both automatic and human evaluation metrics. PMCTG utilizes perturbed masking technology to identify edit locations and reflect the reasonableness of edit actions more intuitively and practically. 

Compared to previous baselines, our approach may either require fewer steps to search for the optimal sentence or equal steps to achieve better results. In this task, our method needs to run only 100 steps while CGMH needs 200 steps for each sample and our method can achieve better results (7.47 vs 7.70 in average NLL). Besides, although G2LC also only needs to run 100 steps for each sample, our method (PMCTG-LSTM) gives better results (7.47 vs 7.56 in average NLL). Although the process requires another BERT model for perturbed masking, we transform a sentence to a batch of vectors and only need to call the BERT model once per search step to calculate the perturbed masking scores for all tokens. Compared to CGMH and UPSA, our method makes full use of each search step to a certain extent, reducing the extra time spent on random strategies. 

Interestingly, PMCTG-LSTM seems to be superior to PMCTG-GPT2 in this task. For one thing, part of the superiority of GPT2 to LSTM is in the semantic richness of the generated sentences. However, in the target dataset, the sentence form and semantics are relatively simple, and therefore the performance of LSTM is comparable to that of GPT2 in cases where there is no need to generate sentences with complex semantics. For another, since keywords are locally ill-formed and semantically distant, the information of keywords may be difficult to support GPT2 to generate reasonable candidates without taking backward probability into account. In contrast, the two-layer LSTM considers both forward and backward probabilities and may be more suitable for generating candidates between two less correlated tokens. 

We find that more keywords may lead to better results, one possible reason is that more keywords can further narrow the search space and facilitate the search of the model.

\noindent \textbf{Case Study.} Some generated examples of PMCTG-LSTM are shown in Table \ref{table:4}. We observe that the proposed model can generate fluent and meaningful sentences while containing the given keywords.

\begin{table}[]
\centering
\resizebox{\columnwidth}{!}{
\begin{tabular}{l|l}
\hline
\textbf{Keywords}                                                     & \textbf{Sentences}                                                                                                      \\ \hline
worried                                                               & \begin{tabular}[c]{@{}l@{}}We are very worried about there .\end{tabular}                                           \\ \hline
agreement                                                               & \begin{tabular}[c]{@{}l@{}}To achieve such an agreement ,\\ it is important .\end{tabular}                                           \\ \hline
\begin{tabular}[c]{@{}l@{}}competition,\\ action\end{tabular}         & \begin{tabular}[c]{@{}l@{}}The shots of competition and\\ action are on display here .\end{tabular}                  \\ \hline
\begin{tabular}[c]{@{}l@{}}change,\\ hours\end{tabular}         & \begin{tabular}[c]{@{}l@{}}This will change it in the next\\ 24 hours .\end{tabular}                  \\ \hline
\begin{tabular}[c]{@{}l@{}}The,greatest,\\  court\end{tabular}        & \begin{tabular}[c]{@{}l@{}}The world's greatest size court\\ will be presented to you .\end{tabular}                \\ \hline
\begin{tabular}[c]{@{}l@{}}I,things,\\  him\end{tabular}        & \begin{tabular}[c]{@{}l@{}}I can do lots of things for him .\end{tabular}                \\ \hline
\begin{tabular}[c]{@{}l@{}}body,\\ advanced,\\ July,funeral\end{tabular} & \begin{tabular}[c]{@{}l@{}}The body was found advanced\\ in July and funeral were held\\ in September .\end{tabular} \\ \hline
\begin{tabular}[c]{@{}l@{}}Miley,more,\\ final,spots\end{tabular}     & \begin{tabular}[c]{@{}l@{}}But Miley Cyrus has played\\  more than three times in\\  the finaltwo spots .\end{tabular} \\ \hline
\end{tabular}}
\caption{Generated examples of PMCTG-LSTM in keywords-to-sentence generation task.}
\label{table:4}
\end{table}

\begin{table}[]
\centering
\begin{tabular}{l|l}
\hline
\textbf{Score}                                                     & \textbf{Description}                                                                                                      \\ \hline
1.00                                                              & \begin{tabular}[c]{@{}l@{}}Two sentences have the completely \\ same meanings.\end{tabular}                                           \\ \hline
0.75                                                              & \begin{tabular}[c]{@{}l@{}}Two sentences have similar meanings \\ with some different details. \end{tabular}                                           \\ \hline
0.50                                                              & \begin{tabular}[c]{@{}l@{}}Two sentences generally have similar \\ meanings with many different details.\end{tabular}                                           \\ \hline
0.25                                                              & \begin{tabular}[c]{@{}l@{}}Two sentences generally have different \\ meanings with some identical details.\end{tabular}                                           \\ \hline
0.00                                                              & \begin{tabular}[c]{@{}l@{}}Two sentences have completely \\different meanings.\end{tabular}                                           \\ \hline

\end{tabular}
\caption{Relevance scoring guideline.}
\label{table:5}
\end{table}

\begin{table*}[]
\centering
\begin{tabular}{l|cccc|cccc}
\hline
 \multirow{2}{*}{\textbf{Models}} &  \multicolumn{4}{c|}{\textbf{Quora}}    & \multicolumn{4}{c}{\textbf{Wikianswer}} \\ 
\cline{2-9}  & iBLEU & BLEU  & R1    & R2    & iBLEU  & BLEU  & R1    & R2\\ 
\hline  ResidualLSTM & 12.67 & 17.57 & 59.22 & 32.40  & 22.94  & 27.36 & 48.52 & 18.71 \\ 
 VAE-SVG-eq        & 15.17 & 20.04 & 59.98 & 33.30  & 26.35  & 32.98 & 50.93 & 19.11 \\
 Pointer-generator & 16.79 & 22.65 & 61.96 & 36.07 & 31.98  & 39.36 & 57.19 & 25.38 \\
 Transformer       & 16.25 & 21.73 & 60.25 & 33.45 & 27.70  & 33.01 & 51.85 & 20.70  \\
 Transformer+Copy  & 17.98 & 24.77 & 63.34 & 37.31 & 31.43  & 37.88 & 55.88 & 23.37 \\
 DNPG              & 18.01 & 25.03 & 67.73 & 37.75 & 34.15  & 41.64 & 57.32 & 25.88 \\ 
\hline 
 Pointer-generator & 5.04  & 6.96  & 41.89 & 12.77 & 21.87  & 27.94 & 53.99 & 20.85 \\
 Transformer+Copy  & 6.17  & 8.15  & 44.89 & 14.79 & 23.25  & 29.22 & 53.33 & 21.02 \\
 Shallow fusion    & 6.04  & 7.95  & 44.87 & 14.79 & 22.57  & 29.76 & 53.54 & 20.68 \\
 MTL               & 4.90   & 6.37  & 37.64 & 11.83 & 18.34 & 23.65 & 48.19 & 17.53 \\
 MTL + Copy        & 7.22  & 9.83  & 47.08 & 19.03 & 21.87  & 30.78 & 54.1  & 21.08 \\
 DNPG              & 10.39 & 16.98 & 56.01 & 28.61 & 25.60  & 35.12 & 56.17 & 23.65 \\ 
\hline 
 VAE               & 8.16  & 13.96 & 44.55 & 22.64 & 17.92  & 24.13 & 31.87 & 12.08 \\
 CGMH              & 9.94  & 15.73 & 48.73 & 26.12 & 20.05  & 26.45 & 43.31 & 16.53 \\
 UPSA              & 12.02 & 18.18 & 56.51 & 30.69 & 24.84  & 32.39 & 54.12 & 21.45 \\
 G2LC-Recognizer   & 14.34 & 20.13 & 58.90 & 32.79 & /      & /     & /     & /     \\
 G2LC-Generator    & 14.46          & 23.27          & \textbf{59.65} & \textbf{33.08}          & /              & /              & /              & /              \\
 PMCTG-LSTM        & 14.79          & 23.73          & 59.21          & 31.92          & 25.66          & 33.87          & 56.21          & 21.92          \\
 PMCTG-GPT2        & \textbf{15.22} & \textbf{24.37} & 59.03          & 32.89 & \textbf{26.13} & \textbf{35.02} & \textbf{56.89} & \textbf{23.21} \\ \hline
\end{tabular}
\caption{Performance on paraphrasing task. R1 and R2 respectively indicate ROUGE1 and ROUGE2. In this table, this first/second/third blocks respectively indicate the results of supervised/domain-adapted supervised/unsupervised methods.  }
\label{table:6}
\end{table*}

\subsection{Paraphrasing}

\noindent \textbf{Experimental Setting.} Paraphrasing aims to convert a sentence to a different surface form but with the same meaning. We evaluate PMCTG on two paraphrase datasets, namely Quora\footnote{http://www.statmt.org/wmt18/translation-task.html} and Wikianswers \cite{DBLP:conf/acl/FaderZE13}. The Quora question pair dataset consists of 140 thousand parallel sentences pairs and 640 thousand non-parallel sentences. The Wikianswers dataset contains 2.3 million question pairs scrawled from the Wikipedia website. We also conduct an experiment on two-layer LSTM (followed as \cite{DBLP:conf/aaai/MiaoZMYL19,DBLP:conf/acl/LiuMMZZS20,DBLP:conf/emnlp/Sha20}) and GPT2 for better comparison. Following previous works \cite{DBLP:conf/acl/LiuMMZZS20} again, we randomly sample 20 thousand sentences respectively in two datasets as test sets and use the other sentences to continually pre-train BERT-based-cased and GPT2 for domain adaption as \cite{DBLP:conf/acl/GururanganMSLBD20}. 

As for hyperparameters, the maximum searching step set in this task is 50 and $\lambda$ are all set as $1$ in equation (13). The initial sentence for searching is the original sentence in the datasets. 

In terms of evaluation metrics, we leverage the representative metrics sentence-level BLEU \cite{DBLP:conf/acl/PapineniRWZ02} and ROUGE \cite{lin2004rouge} as the basic metrics. In addition, as stated in \cite{DBLP:conf/acl/SunZ12}, standard BLEU and ROUGE could not reflect the diversity between the generated and original sentences. Therefore, we adopt iBLEU \cite{DBLP:conf/acl/SunZ12} which penalizes the generated sentences with high similarity with the original ones as an additional evaluation metric. Besides, we also invite experts to evaluate the generated paraphrases. Specifically, we sample 300 sentences from the Quora test set and ask 3 experts to score each sentence according to two aspects: relevance and fluency. The evaluation criterion is again consistent with the previous works \cite{DBLP:conf/aaai/MiaoZMYL19,DBLP:conf/acl/LiuMMZZS20}. The scoring guidelines are shown in Table \ref{table:2} and Table \ref{table:5}.

\noindent \textbf{Baseline.} We compare our methods with three types of baseline:

\begin{itemize}
\setlength{\itemsep}{0pt}
\setlength{\parsep}{0pt}
\setlength{\parskip}{0pt}
    \item \textbf{Supervised methods} are original sequence-to-sequence models trained in in-domain supervised data, including ResidualLSTM \cite{DBLP:conf/coling/PrakashHLDQLF16}, VAE-SVG-eq \cite{DBLP:conf/aaai/GuptaASR18}, Pointer-generator \cite{DBLP:conf/acl/SeeLM17}, the Transformer \cite{DBLP:conf/nips/VaswaniSPUJGKP17}, and DNPG (the decomposable neural paraphrase generation) \cite{DBLP:conf/acl/LiJSL19}.
    \item \textbf{Domain-adapted supervised methods} train models in one domain and then adapt them to another domain, including shallow fusion \cite{DBLP:journals/corr/GulcehreFXCBLBS15} and multi-task learning (MTL) method \cite{DBLP:conf/emnlp/DomhanH17}.
    \item \textbf{Unsupervised methods} that are free of any supervised data and easily adapted to multiple new domains, including VAE \cite{DBLP:journals/corr/KingmaW13}, CGMH \cite{DBLP:conf/aaai/MiaoZMYL19}, UPSA \cite{DBLP:conf/acl/LiuMMZZS20}, and the recurrent state-of-the-art method G2LC \cite{DBLP:conf/emnlp/Sha20}. Notably, G2LC has two variants of G2LC-Generator and G2LC-Recognizer.
\end{itemize}

\noindent \textbf{Automatic Evaluation Results.} Table \ref{table:6} presents the results of multiple methods on the paraphrasing task. From the first part of Table \ref{table:6}, we can see that supervised methods significantly outperform the other two kinds of methods. The supervised models were trained on 100 thousand question pairs for Quora and 500 thousand question pairs for Wikianswers. Their superiority indicates the effectiveness of learning knowledge from massive parallel data. However, such in-domain supervised data is hard to obtain in real-world applications.

Besides, the second section of Table \ref{table:6} shows the domain-adapted supervised models' performance. These models are trained in one domain (Quora or Wikianswers) and then evaluated in another domain (Wikianswers or Quora). Their performances are much lower than in-domain supervised models' performances. This demonstrates the poor generalizability of supervised models and calls for the need for unsupervised methods.

The last section of Table \ref{table:6} shows the results of multiple unsupervised methods. VAE seems to work worst on both datasets, which suggests that paraphrasing by latent space sampling performs not as well as local edit methods. PMCTG achieves the best performance in most cases, which indicates the effectiveness of PMCTG again. Unsupervised PMCTG does not require parallel data and can easily generalize to new domains, thus some unsupervised methods tend to achieve higher performance than the domain-adapted supervised models. In addition, it is worthwhile to note that the performance of some unsupervised methods (UPSA, G2LC, and PMCTG) is even better than some supervised methods (Residual LSTM and VAE-SVG-eq), which indicates that the gap between supervised and unsupervised methods has narrowed due to the effective searching strategies of the local edit-based methods. In addition, different from the keywords-to-sentence generation task, GPT2 works better than two-layer LSTM in the paraphrasing task. We believe that given a partially fluent text, GPT2 can generate more reasonable candidates due to its powerful language modeling capability.

\begin{table}[]
\centering
\begin{tabular}{l|c|c}
\hline
\textbf{Method}  & \textbf{Relevance} & \textbf{Fluency} \\ \hline
VAE              & 0.53               & 0.64             \\
CGMH             & 0.62               & 0.70             \\
UPSA             & 0.75               & 0.73             \\
G2LC(Recognizer) & 0.79               & 0.77             \\
G2LC(Generator)  & \textbf{0.81}      & 0.78             \\ \hline
PMCTG-GPT2       & 0.76               & \textbf{0.81}    \\ \hline
\end{tabular}
\caption{Human evaluation results on paraphrasing.}
\label{table:7}
\end{table}

\noindent \textbf{Human Evaluation Results.} From Table \ref{table:7}, we show PMCTG-GPT2 achieves state-of-the-art performance in terms of fluency, but still suffers from relevance. We plan to improve its relevance in future research.

\begin{table}[]
\centering
\resizebox{\columnwidth}{!}{
\begin{tabular}{l|l}
\hline
\textbf{Type} & \textbf{Sentence}                                                                                               \\ \hline\hline
Ori           & \begin{tabular}[c]{@{}l@{}}what can make physics easy to learn?\end{tabular}              \\\hline
Gen           & \begin{tabular}[c]{@{}l@{}}how to learn physics easily?\end{tabular}                     \\\hline

Ref           & \begin{tabular}[c]{@{}l@{}}how can you make physics easy to learn?\end{tabular} \\ \hline
\hline
Ori           & \begin{tabular}[c]{@{}l@{}}is it possible to pursue many different things\\ in life?\end{tabular}              \\\hline
Gen           & \begin{tabular}[c]{@{}l@{}}is it good to buy many different things in life?\end{tabular}                     \\\hline
Ref           & \begin{tabular}[c]{@{}l@{}}how do i refuse to choose between different\\ things to do in my life?\end{tabular} \\ \hline
\hline
Ori           & \begin{tabular}[c]{@{}l@{}}how do i choose a journal to publish my paper?\end{tabular}                       \\\hline
Gen           & \begin{tabular}[c]{@{}l@{}}how do you choose a journal to publish your\\ first book?\end{tabular}              \\\hline
Ref           & where do i publish my paper?                                                                                    \\\hline \hline
Ori           & \begin{tabular}[c]{@{}l@{}}where can i get free books to read or download?\end{tabular}                      \\\hline
Gen           & \begin{tabular}[c]{@{}l@{}}where did i download free books to read?\end{tabular}                             \\\hline
Ref           & where can i get free books?                                                                                     \\\hline \hline
\end{tabular}}
\caption{Generated examples of PMCTG-GPT2 in paraphrasing task.}
\label{table:8}
\end{table}

\noindent \textbf{Case Study.} Table \ref{table:8} lists some representative generated examples from PMCTG-GPT2. They show the four most common types of paraphrasing for the proposed method. The first type is the change of syntax such as the interchange of “\textit{what can…}” and “\textit{how to…}” as in the first example. The second type is the change of adjective such as the second example where the “\textit{possible}” is changed into “\textit{good}”. The third type is the change of personal pronouns such as the interchange of “\textit{you}” and “\textit{I}” in the third example. The last type is the change of tense, the most common is the interchange of general past tense and general present tense as the last example. In general, one limitation of the proposed model is the relatively low expressive diversity of generated sentences. One possible reason is that since each search step modifies only one token, and the unit of conversion from one expression to another is usually phrases or sentence blocks, thus the model may be biased not to search in that direction.

\section{Conclusion}
We propose a method PMCTG to improve the previous stochastic searching methods in the topic of unsupervised constrained generation. PMCTG leverages perturbed masking technique to find the best edit position and leverages newly designed multiple scoring functions to decide the best edit action. We evaluate the proposed method on two representative tasks: keywords-to-sentence generation (hard constraints) and paraphrasing (soft constraints). Experimental results demonstrate the effectiveness of the proposed method which achieves competitive results on three datasets over multiple advanced baseline methods. We plan to improve the diversity and relevance of the generated sentences in future work.

\section{Acknowledgement}
We are grateful for the inspiration and project support from "The World 3" in NetEase Games. We also thank anonymous reviewers for their comments and suggestions.

\bibliography{reference}
\bibliographystyle{acl_natbib}

\end{document}